\documentclass[sigconf, review=false, screen=true, authordraft=false]{acmart}

\AtBeginDocument{%
  \providecommand\BibTeX{{%
    \normalfont B\kern-0.5em{\scshape i\kern-0.25em b}\kern-0.8em\TeX}}}

\setcopyright{acmlicensed}
\copyrightyear{2024}
\acmYear{2024}
\acmDOI{XXXXXXX.XXXXXXX}

\acmConference[KDD '24]{KDD}{August 25-29, 2024}{Barcelona, Spain}
\acmSubmissionID{444}

\begin{document}

\title{Video Annotator: A framework for efficiently building video classifiers using vision-language models and active learning}



\author{Amir Ziai}
\email{aziai@netflix.com}
\affiliation{%
  \institution{Netflix Inc}
  \city{Los Gatos}
  \state{CA}
  \country{USA}
}

\author{Aneesh Vartakavi}
\email{avartakavi@netflix.com}
\affiliation{%
  \institution{Netflix Inc}
  \city{Los Gatos}
  \state{CA}
  \country{USA}
}

\begin{abstract}

High-quality and consistent annotations are fundamental to the successful development of robust machine learning models. Traditional data annotation methods are resource-intensive and inefficient, often leading to a reliance on third-party annotators who are not the domain experts. Hard samples, which are usually the most informative for model training, tend to be difficult to label accurately and consistently without business context. These can arise unpredictably during the annotation process, requiring a variable number of iterations and rounds of feedback, leading to unforeseen expenses and time commitments to guarantee quality.

We posit that more direct involvement of domain experts, using a human-in-the-loop system, can resolve many of these practical challenges. We propose a novel framework we call Video Annotator (VA) for annotating, managing, and iterating on video classification datasets. Our approach offers a new paradigm for an end-user-centered model development process, enhancing the efficiency, usability, and effectiveness of video classifiers. Uniquely, VA allows for a continuous annotation process, seamlessly integrating data collection and model training.

We leverage the zero-shot capabilities of vision-language foundation models combined with active learning techniques, and demonstrate that VA enables the efficient creation of high-quality models. VA achieves a median 8.3 point improvement in Average Precision relative to the most competitive baseline across a wide-ranging assortment of tasks. We release a dataset with 153k labels across 56 video understanding tasks annotated by three professional video editors using VA, and also release code to replicate our experiments at \href{http://github.com/netflix/videoannotator}{\textcolor{blue}{github.com/netflix/videoannotator}}.




\end{abstract}




\maketitle

\section{Introduction}
\label{sec:intro}

Conventional techniques for training machine learning classifiers are resource intensive. They involve a cycle where domain experts annotate a dataset, which is then transferred to data scientists to train models, review outcomes, and make changes. This labeling process tends to be time-consuming and inefficient, often halting after a few annotation cycles. Consequently, less effort is invested in annotating larger datasets compared to iterating on more complex models and algorithmic methods to improve performance and fix edge cases, as a result of which ML systems grow rapidly in complexity. Furthermore, constraints on time and resources often result in leveraging third-party annotators rather than domain experts. These annotators perform the labeling task without a deep understanding of the model's intended deployment or usage, often making consistent labeling of borderline or hard examples, especially in more subjective tasks, a challenge. This often necessitates multiple review rounds with domain experts, leading to unexpected costs and delays. This lengthy cycle can also result in model drift, as it takes longer to fix edge cases and deploy new models, potentially hurting usefulness and stakeholder trust in these technologies.

We suggest that more direct involvement of domain experts, using a human-in-the-loop system, can resolve many of these practical challenges described above. We introduce a novel framework, Video Annotator (VA), for annotating, managing, and iterating on video classification datasets. Our interactive framework employs active learning techniques, to guide users to focus their efforts on progressively harder examples, enhancing the model's sample efficiency and keeping costs low. Equipped with sufficient knowledge and context, they can rapidly make informed decisions on hard samples during the annotation process. We leverage the zero-shot capabilities of large vision-language models to aid the annotation process and to serve as backbones for lightweight binary classifiers that we train for each label. This design simplifies model deployment, enabling us to scale to a large number of labels without significantly increasing complexity. VA seamlessly integrates model building into the data annotation process, facilitating user validation of the model before deployment, therefore helping with building trust and fostering a sense of ownership. VA supports a continuous annotation process, allowing users to rapidly deploy models, monitor their quality in production, and swiftly fix any edge cases by annotating a few more examples and deploying a new model version. This self-service architecture empowers users to make improvements without active involvement of data scientists or third-party annotators, allowing for fast iteration. 

\begin{figure}
    \centering
    \includegraphics[width=1.0\linewidth]{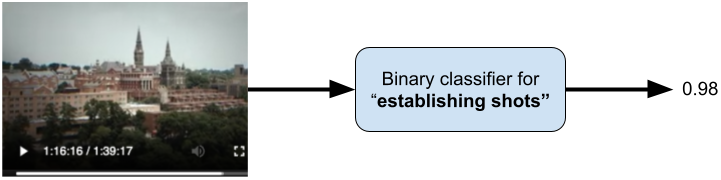}
    \caption[]{Functional view of a binary video classifier. A few-second clip from "Operation Varsity Blues: The College Admissions Scandal" \cite{operation_varsity_blues} is passed to a binary classifier for detecting the "establishing shots" \cite{mittal2007detecting} label. The classifier outputs a very high score, indicating that the video clip is very likely an establishing shot.}
    \label{fig:vid_clf}
\end{figure}

We design VA to assist in granular video understanding \cite{huang2018makes,diba2020large} which requires the identification of visuals, concepts, and events within video segments. Video understanding is fundamental for numerous applications such as search and discovery \cite{ava_discovery,mup,building_in_video_search,scene_change}, personalization \cite{artwork_personalization,new_series,dynamic_sizzles}, and the creation of promotional assets \cite{chen2023match,new_series,creative_insights}. Some forms of video understanding requires knowledge of a broader context or narrative, for example, identifying the "inciting incidents" or story turning points \cite{papalampidi2019movie}. Other forms are relatively "context-free", for example determining whether a clip is set during the day or night is independent of the other parts of the video. We focus on "context-free" video classification in our work, which is the task of assigning a label to an arbitrary-length video clip without broader context, as illustrated in Fig \ref{fig:vid_clf}.

In summary, our main contributions are:
\begin{itemize}
\item We present Video Annotator (VA), a novel framework to build video classifiers efficiently using pre-trained visual-language models, active learning, and direct involvement of domain experts in the training process.
\item We deployed VA internally at Netflix and found applicability in a variety of business problems. In this work, we conduct experiments that demonstrate that our system enjoys improved sample efficiency, visual diversity, and model quality, with a 8.3 median point improvement in Average Precision (AP) \cite{ranking_ap} relative to the most competitive baseline.
\item We release a dataset with 153k labels across 56 video understanding tasks annotated by three professional video editors using VA, and also release code to replicate our experiments.
\end{itemize}

\section{Related work}

\subsection{Vision Language Models (VLMs)}
Self-supervised methods using convolutional or transformer-based architectures have dominated much of the progress in video understanding in recent years \cite{radford2021learning,luo2022clip4clip,alayrac2022flamingo,bao2021beit,liu2022video,li2022blip,khan2022transformers,zhang2021vinvl}. Specifically, VLMs such as CLIP \cite{radford2021learning}, demonstrate very good zero-shot natural language understanding performance across a wide variety of visual tasks \cite{luo2022clip4clip,chen2023valor,arnab2021vivit,li2023unmasked,xue2022clip}. However, zero-shot methods often lack the requisite nuance and reliability in many real-world applications \cite{wang2023review}, especially for the long tail of visual concepts \cite{moll2023seesaw}\cite{romero2023zelda}.

SeeSaw \cite{moll2023seesaw} tackles this problem by aligning CLIP \cite{radford2021learning} query vectors using user feedback via an interactive system for search over image databases. Similar to SeeSaw, we use VLMs to enable search. In contrast to SeeSaw, our approach focuses on video classification for the purpose of generalization to unseen videos vs. interactive search sessions, and uses active learning instead of query alignment.

\subsection{Video Annotation Tooling}

Numerous commercial and open-source tools for video analytics and annotation exist \cite{zhang2021survey,romero2023zelda,alsaid2022datascope,shrestha2023feva}. Many of these approaches focus on identifying and tracking specific objects within a video \cite{stanchev2020automating,kim2013semi,ashangani2016semantic}, while others focus on interactive search sessions \cite{romero2023zelda,moll2023seesaw,daum2023vocalexplore}. In contrast, our approach focuses on video classification for highly specific tasks. Our goal is to construct models that can generalize to unseen data and can be used by either end-users or other algorithms, rather than focusing on interactive search sessions over a fixed dataset. Nevertheless, we outline some approaches that share similarities with our work.

Zelda \cite{romero2023zelda} is a video analytics tool that employs VLMs to assist users in retrieving relevant and diverse results through prompt engineering techniques. Similar to Zelda, we use VLMs to facilitate semantic search using natural language. In contrast to Zelda, we leverage active learning to annotate a labeled dataset that captures nuances that are hard to express solely with prompt engineering.

VOCALExplore \cite{daum2023vocalexplore} offers a system for early data exploration on video datasets using active learning. Both approaches harness VLMs, but VOCALExplore uses them only as feature extractors, while we leverage VLMs as both feature extractors and to enable text-video search. While both VA and VOCALExplore are interactive systems aiming to minimize latency, they adopt different strategies to achieve this goal. VA employs a pre-processing step over a large corpus of videos, enabling both text-to-video search and uses lightweight models over a VLM backbone to reduce latency. In contrast, VOCALExplore uses a Task Scheduler and active learning strategies to avoid the pre-processing step. VA is less restrictive and offers more user choice and freedom, allowing users to freely search for specific examples that they are interested in labeling, while VOCALExplore requires users to specify a labeling budget and suggests a batch of samples to review.

To the best of our knowledge, VA represents the first-of-its-kind system for granular video classification, leveraging VLMs to facilitate semantic search through natural language and active learning to enhance sample efficiency. It offers a unique approach to annotating, managing, and iterating on video classification datasets, emphasizing the importance of direct involvement of domain experts in a human-in-the-loop system.

\section{Methodology}
\label{methodology}

\begin{figure}
    \centering
    \includegraphics[width=1.0\linewidth]{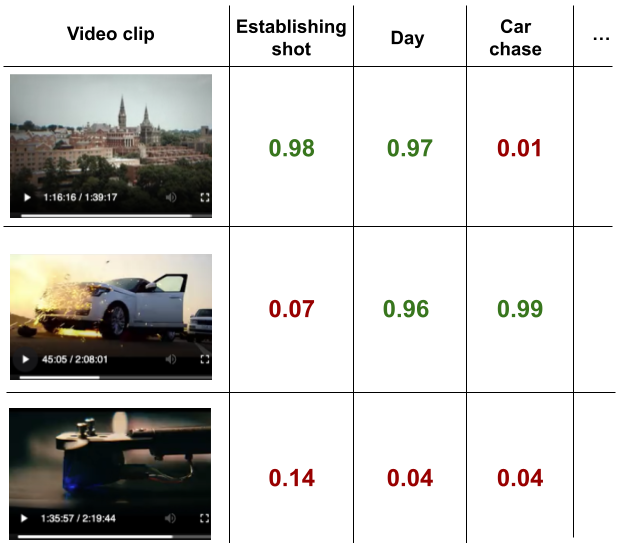}
    \caption[]{
    Three video clips and the corresponding binary classifier scores for three video understanding labels. Note that these labels are not mutually exclusive. Video clips are from Operation Varsity Blues: The College Admissions Scandal \cite{operation_varsity_blues}, 6 Underground \cite{6_underground}, and Leave The World Behind \cite{leave_the_world_behind}, respectively.}
    \label{fig:vid_und}
\end{figure}

\begin{figure*}
    \centering
    \includegraphics[width=0.85\linewidth]{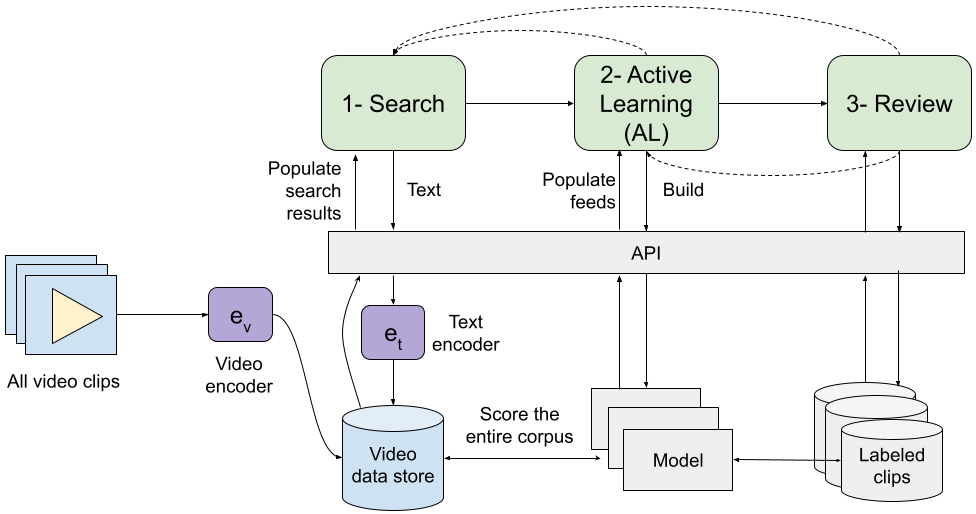}
    \caption{
    The main annotation process of Video Annotator (VA) involves three steps, illustrated by the green components. In step 1, the annotator retrieves video clips from text queries which are encoded using $e_t$. In Step 2, the clips are labeled and used to build a classifier for further refinement. Finally, step 3 is an opportunity to review all annotated clips. Note that this process is rarely linear, and the annotator can easily navigate between steps.
    }
    \label{fig:system}
\end{figure*}

We describe our proposed framework, called Video Annotator (VA), which is designed for building binary video classifiers for an extensible set of labels (see Fig \ref{fig:vid_und}), enabling granular video understanding. This formulation allows for improving one model independent of the others. Fig \ref{fig:system} depicts an overview of VA. Our system uses video ($e_v$) and text encoders ($e_t$) from a Vision-Language Model (VLM) to extract embeddings. We gather a diverse set of video clips, the embeddings for which are extracted using $e_v$, and stored for efficient retrieval (e.g. FAISS \cite{johnson2019billion}) alongside other metadata about each clip. Users can begin building classifiers after this pre-processing step is complete.

\begin{figure}
    \centering
    \includegraphics[width=0.9\linewidth]{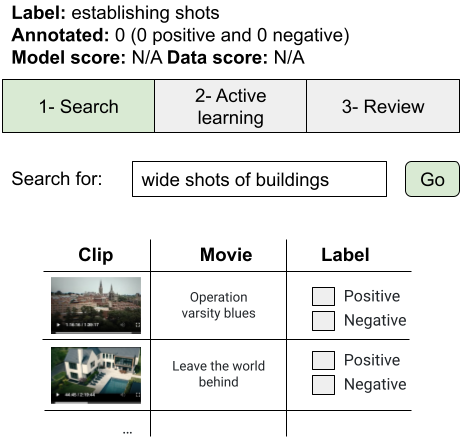}
    \caption[]{Step 1: Text-to-video search to bootstrap the annotation process.}
    \label{fig:step1}
\end{figure}

Users begin by finding an initial set of examples to bootstrap the annotation process. We leverage text-to-video search to enable this, using the text encoder ($e_t$) of a VLM. For example, an annotator working on the "establishing shots" \cite{mittal2007detecting} label may start the process by searching for "wide shots of buildings". In film-making, an establishing shot is a wide shot (i.e. video clip between two consecutive cuts) of a building or a landscape that is intended for establishing the time and location of the scene. We also allow users to search using other available fields (e.g. clips from a specific movie). Fig \ref{fig:step1} illustrates this example. Annotators are encouraged to label as many diverse examples as possible, and VA requires at least 10 positive and 10 negative annotations before the annotator can proceed to the next stage.

\begin{figure}
    \centering
    \includegraphics[width=0.9\linewidth]{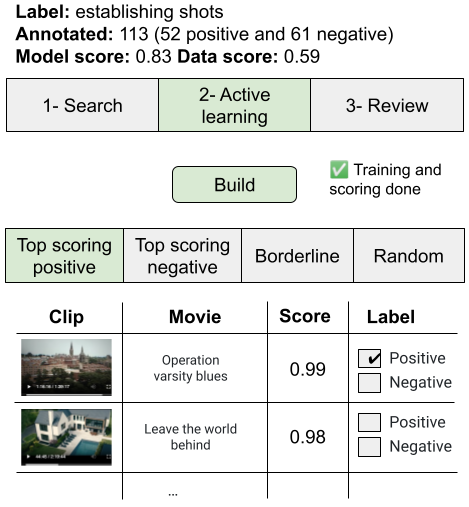}
    \caption[]{Step 2 - Active Learning (AL) loop. The annotator clicks on build, which initiates classifier training and scoring of all clips in $C$. Scored clips are displayed in four feeds. }
    \label{fig:step2}
\end{figure}

The next stage involves a classic Active Learning (AL) loop \cite{settles2009active}. VA builds a binary classifier over the video embeddings, which is subsequently used to score all clips in the corpus. We choose lightweight classifiers (e.g. logistic regression) to minimize latency, so that users only have to wait for a few seconds for the results. Scored clips are presented in four feeds (i.e. ordered list of clips) for further annotation and refinement; top-scoring positive, top-scoring negative, borderline (i.e. low confidence), and random, as illustrated in Fig \ref{fig:step2}. The annotator can label additional clips in any of the four feeds and build a new classifier, repeating the AL process for as many iterations as desired. We also include a view of all the labeled examples so far, allowing users to review all their annotations in a single place. VA keeps a history of all annotated clips and the corresponding classifiers. 

The top-scoring positive and negative feeds display examples with the highest and lowest scores respectively. Our users reported that this provided a valuable indication as to whether the classifier has picked up the correct concepts in the early stages of training, since clips that have already been annotated tend to appear in these feeds. Our users were also able to spot cases where a model was learning incorrect concept due to a bias in the training data that they were able to subsequently fix. For example, if all positive instances occur during the day and negatives at night, the model may mistakenly learn to differentiate between day and night, rather than accurately identifying the intended label.

The third feed contains examples that the model is not confident about. We use a simple uncertainty sampling method, the absolute deviation of the model score from 0.5, which requires the use of calibrated models \cite{cohen2004properties}. This feed helps with discovering interesting edge cases and inspires the need for labeling additional concepts. For instance, the annotator notices that the model is confused about animated content, and can find and label additional examples using visual search.

The fourth feed consists of randomly selected clips which helps to annotate diverse examples. Our early experiments suggested that only focusing on search and uncertainty sampling may lead to biased datasets that generalize poorly to unseen clips.

\subsection{Metrics}

VA guides the annotator through the labeling process by displaying model quality and data diversity scores. 

\subsubsection{Model quality score}
We want the classifier to reliably separate positives from negatives. To measure this, we use $K$-fold cross-validation and report the 25th percentile across the $K$ Balanced Accuracy (BA) \cite{brodersen2010balanced} scores. We picked BA because it's not sensitive to class imbalance and is an intuitive metric to understand. We report the 25th percentile instead of mean/median in order to be more conservative in our reporting. We use $K=5$ in this work.

\subsubsection{Data diversity score}
\label{section_data_diversity_score}

A model solely optimized for quality could potentially exhibit bias and fail to generalize. A diverse training set encompassing a broad spectrum of visual concepts is necessary for the construction of robust models. We therefore formulated a heuristic we call the data diversity score to guide our users. 

\begin{figure}
    \centering
    \includegraphics[width=0.9\linewidth]{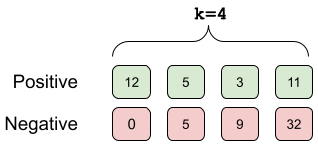}
    \caption{Data score calculation for a simulated dataset with 77 annotations (31 positive and 46 negative) and $k=4$. Each element is capped to a maximum value $s=10$, summed, and divided by $2 \times k \times s = 80$. In this case, the data score is $(10 + 5 + 3 + 10 + 0 + 5 + 9 + 10)/(2 \times 4 \times 10) = 0.65$. }
    \label{fig:data_score_calc}
\end{figure}

The goal is to accumulate annotations from a variety of visual elements. We identify these elements using $k$-means clustering over the embeddings of all the clips within the corpus. We map the positive and negative annotations independently onto these clusters, and compute the resulting counts, capping it to a maximum value $s$. We then sum all the counts, and normalize (i.e. dividing by $2 \times k \times s$). An example is depicted in Figure \ref{fig:data_score_calc}. We use $k=s=10$ for this study. The scores range from 0 to 1. Labeling examples from the randomly selected feed helps improve this score, but note that it may be difficult to achieve a perfect score. In the worst case, the best achievable score is 0.5, which happens when a label perfectly aligns with one of the clusters.


\section{Data}
\label{data}

We used VA to collect ~153K annotations across 56 labels that cover a variety of tasks in video understanding, categorized into 8 groups as shown in Table \ref{table:labels}. We used the Condensed Movies Dataset \cite{bain2020condensed}, which consists of close to 30k movie scenes belonging to 3.5k movies across a wide range of genres, decades, and countries.  First, we used PySceneDetect \cite{pyscene} to segment each scene into individual shots. Then, we extracted shot-level Clip4CLIP (C4C) \cite{luo2022clip4clip} embeddings and used those to remove near-duplicate shots. We followed the embedding extraction and de-duplication approach in \cite{chen2023match}, which yielded a final set of ~500k shots that we use as our video corpus. The average shot duration for this final set is 3.4 seconds. The 10th and 90th percentile shot durations are 0.8 and 6.5 seconds respectively.

We utilized the skills of three professional video editors, all of whom possess a deep understanding of film and video editing techniques. We conducted an initial hour-long training session on the usage of VA. We shared a text description and a few relevant keywords for each label, along with sample positive and negative clips. We collected three annotations for each label. Since VA allows the users the freedom to choose what clips to annotate, it is unlikely that all the users would label the same examples. We therefore randomly selected a "primary annotator" for each label who would complete the task first. The remaining "secondary annotators" would independently label the set of examples labeled by the primary annotator. We instructed the primary annotators to label at least 100 positive and negative annotations each, and stop when they achieved model and data scores of at least 80, or they had annotated 1K total examples. We found high agreement among annotators for most labels. Agreement \cite{kappas}, defined as the percentage of annotations where all 3 annotators picked exactly the same label, has min, max, and average of 62\%, 99\%, and 84\% respectively.

To benchmark the performance of VA against some baselines, we also collected data using two other strategies: random sampling and zero-shot retrieval. To maintain consistency, we had the same three editors label these examples independently. They conducted this task without using the VA interface and skipped the samples already labeled using VA. For the random strategy, we sampled 1K random shots and had these labeled independently for each label (i.e. the set is shared across labels). For zero-shot retrieval, we employed C4C \cite{luo2022clip4clip} to extract the top 1K clips using text-to-video retrieval, using the label text as the query.

\begin{table}[htbp]
\centering
\begin{tabular}{@{}ccp{4.9cm}@{}}
\toprule
\textbf{Group}   & \textbf{\#} & \multicolumn{1}{c}{\textbf{Labels}} \\ \midrule
Emotions         & 5           & anger, happy, laughter, sad, scared \\ \midrule
Events / actions & 4           & car chase, fight, interview, run    \\ \midrule
Focus         & 3           & animal, character, object \\ \midrule
Genres         & 6           & action, drama, fantasy, horror, romance, sci-fi \\ \midrule
Motion         & 6           & handheld, jump-scare, pan, slow-motion, timelapse, zoom \\ \midrule
Sensitivities         & 7           & alcohol, drugs, gore, intimacy, nudity, smoking, violence \\ \midrule
Shot types         & 22           & aerial, closeup, cowboy-shot, dutch-angle, establishing-shots, extreme-close-up, extreme-wide-shot, eye-level, group-shot, high-angle, insert-shot, low-angle, medium, over-the-shoulder-shot, overhead-shot, point-of-view-shot, shutter-shot, single-shot, static-shot, tilt-shot, two-shot, wide \\ \midrule
Time / location         & 3           & day, golden-hour, interior \\ \bottomrule
\end{tabular}
\caption{Annotated labels categorized into 8 groups. The second column is the number of labels for each group.}
\label{table:labels}
\end{table}


\section{Experiments}

This section details our experiments to study the sample efficiency of VA compared to baselines. For every label, 20\% of the labeled data, obtained through uniform sampling, is allocated for testing, while the remainder serves as the training set. Each experiment is repeated five times, each time with independently sampled bootstraps of the training set and evaluated against the same test set. The evaluation metric employed is the mean Average Precision ($AP$) \cite{ranking_ap}. We used the majority vote among the three annotators as the ground truth binary label for $AP$ computation.



\subsection{Experiment 1: Sample Efficiency}
\label{exp1}
We compare the sample efficiency of VA to the following methods by evaluating performance at different training set sizes (denoted by $n$):

\begin{itemize}
\item \textbf{Method 1: Baseline (B)}: We use expected $AP$ as the baseline, which corresponds to $AP$ for random ranking \cite{ranking_ap,ranking}.

\item \textbf{Method 2: Zero Shot (ZS)}: Using the C4C\cite{luo2022clip4clip} model, we use the cosine similarity between the embedding of each clip and the text representation of the label, which we use for computing AP.

\item \textbf{Method 3: Zero Shot Classifier (ZSC)}: We train a binary classifier using the top-$n$ annotated zero-shot clips (see Section \ref{data} for details), sorted in descending order by cosine similarity.

\item \textbf{Method 4: Randomly-selected Clip Classifier (RCC)}: We train a binary classifier using the first $n$ annotated random clips (see Section \ref{data} for details).

\item \textbf{Method 5: Combined classifier (CC)}: We use a binary classifier trained with 50\% of the data selected from zero-shot and the other 50\% from random clips.
\end{itemize}

In instances where the number of positive samples is extremely limited, random sampling or zero-shot retrieval may not yield sufficient positive or negative clips for model training \cite{daum2023vocalexplore}. This problem tends to be more pronounced for smaller values of $n$. We detail this issue in the next section.

\subsubsection{Coverage}
\label{coverage}
\begin{figure}
    \centering
    \includegraphics[width=1.0\linewidth]{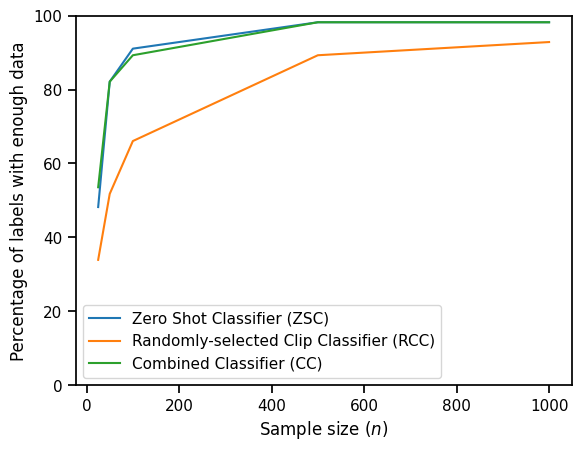}
    \caption[]{Percentage of labels with at least $K$ positive and negative labels for training a classifier, as a function of $n$. With smaller values of $n$, random sampling is unlikely to retrieve enough positive clips for labels with low positive rate. }
    \label{fig:cov}
\end{figure}

As the sample size decreases, the likelihood of finding adequate positive samples to train with also drops. Since we use $K$-fold cross-validation with $K=5$, we require a minimum of 5 positive and 5 negative samples. Fig \ref{fig:cov} illustrates the percentage of labels for which a classifier can be trained as a function of sample size. For example, with 25 samples (i.e. $n=25$), only 34\% of labels yield enough positives when clips are randomly selected. Similarly, only 50\% yield sufficient positive and negative examples when using zero-shot retrieval.

Since models cannot be trained for these cases, we can either impute by zero, which will bring down the aggregate performance for the impacted methods, or exclude them. We chose the latter, and exclude labels without a minimum of $K$ positive or negative annotations for each value of $n$ from further analysis. Note that VA, by design, ensures the selection of a minimum quantity of positives and negatives by the annotator.





\subsubsection{Model quality}
\label{model_quality}

\begin{figure}
    \centering
    \includegraphics[width=1.0\linewidth]{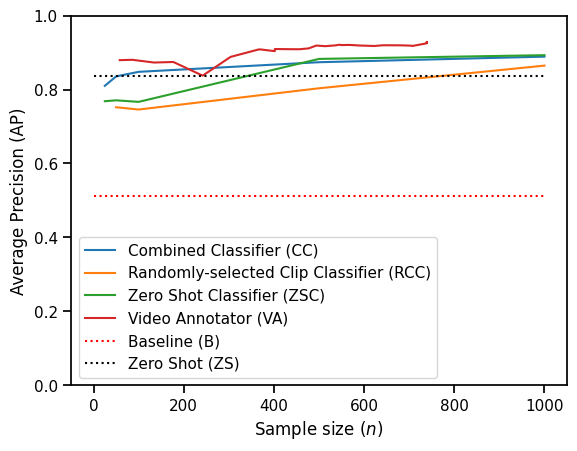}
    \caption[]{Model quality (i.e. AP) as a function of $n$, for the "establishing shots" label. We observe that all methods outperform the baseline, and that all methods benefit from additional annotated data, albeit to varying degrees. }
    \label{fig:est}
\end{figure}

\begin{figure}
    \centering
    \includegraphics[width=1.0\linewidth]{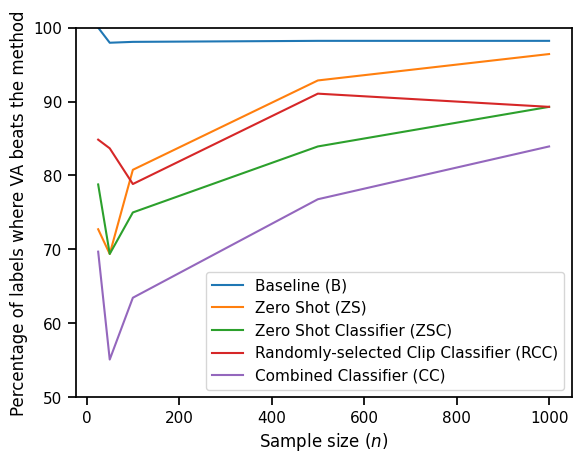}
    \caption[]{Percentage of labels where VA beats other methods as a function of $n$. We see that VA outperforms all other methods in the majority of cases across all values of $n$. }
    \label{fig:exp1_win}
\end{figure}

In this section, we examine model quality, quantified by Average Precision (AP), as a function of sample size $n$. We illustrate the values for the "establishing shot" labels in Fig \ref{fig:est}. We aggregate performance over all labels in Fig \ref{fig:exp1_win}. We find that all methods outperform the baseline, and VA surpasses other methods for a majority of labels across all sample sizes. As discussed in Section \ref{coverage}, the performance of ZS, ZSC, and RCC for smaller values of $n$ are inflated due to the removal of labels with insufficient positive and negative samples.

We also find that the improvement in model quality varies across label groups. Table \ref{table:lift_above_cc} details the median AP gain for VA in comparison to CC across varying $n$ values, with CC selected due to its superior performance among other methods. The gain for each label is calculated as follows: $(AP_{\text{VA}} - AP_{\text{CC}}) \times 100 $.
 
The results from this experiment suggest that VA, in addition to enhanced sample efficiency, offers continuous improvements in model quality as the number of samples increases. 

\begin{table}[]
\centering
\begin{tabular}{@{}lllll@{}}
\toprule
\multicolumn{1}{c}{\textbf{Group}} & \multicolumn{1}{c}{\textbf{n=25}} & \multicolumn{1}{c}{\textbf{n=50}} & \multicolumn{1}{c}{\textbf{n=100}} & \multicolumn{1}{c}{\textbf{n=1000}} \\ \midrule
Emotions                           &  -6.0 &  -5.1 &  -4.2 &   3.7 \\ \midrule
Events / actions                   & 6.9 &   5.5 &   5.7 &   3.8 \\ \midrule
Focus                              & 1.2 &   -1.2 &   -0.4 &   1.8 \\ \midrule
Genres                             & 6.6 &   5.3 &   3.9 &   1.8 \\ \midrule
Motion                             & 31.5 &   1.2 &   5.6 &   3.0 \\ \midrule
Sensitivities                      & 0.5 &  -0.3 &   0.8 &   3.4 \\ \midrule
Shot types                         & 1.2 &  0.6 &   0.5 &   2.3 \\ \midrule
Time / location                    & 4.6 &  -8.5 &  1.3 &   1.4 \\ \bottomrule
\textbf{Overall}                   & \textbf{1.2} &  \textbf{0.4} &  \textbf{1.5} &   \textbf{2.9} \\ \bottomrule
\end{tabular}
\caption[]{Model quality gain for using VA relative to Combined Classifier (CC). For each group, we take the median gain across all labels belonging to the group and a specific value of $n$. The last row represents the median across all labels.}
\label{table:lift_above_cc}
\end{table}


\subsubsection{Data diversity score}
\begin{figure}
    \centering
    \includegraphics[width=1.0\linewidth]{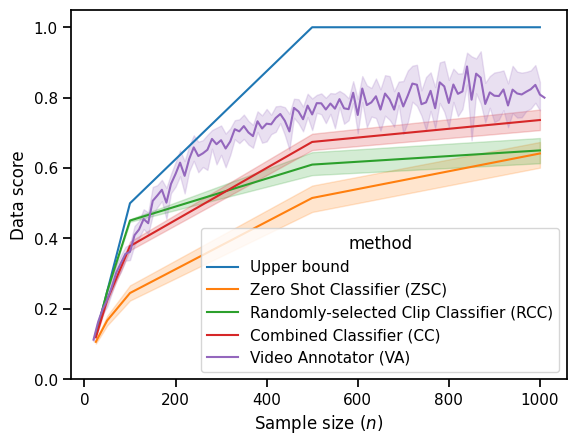}
    \caption[]{Mean and standard deviation of the data score for different methods as a function of $n$. Upper bound is the maximum possible score at each value $n$. Data score for all methods except for VA are computed at fixed values $n \in \{25, 50, 100, 500, 1000\}$.}
    \label{fig:exp1_data_score}
\end{figure}

As referenced in Section \ref{section_data_diversity_score}, VA motivates annotators to label visually diverse clips. Fig \ref{fig:exp1_data_score} represents the aggregated data diversity score as a function of $n$. We find that VA produces datasets with higher visual diversity relative to the other methods, which possibly contributes to the improved generalization performance we noted earlier.

\subsection{Experiment 2: annotation candidate source selection}
\label{exp2}

VA offers annotators the flexibility to choose the source of candidates for annotation, providing options between search and any of the four feeds (as discussed in Section \ref{methodology}). As demonstrated in Experiment 1, this strategy typically yields superior model quality and data diversity scores.



Nevertheless, anecdotal feedback suggested that our users, while appreciating the VA workflow and user experience, were sometimes uncertain about the optimal action to take following a model retraining step. VA doesn't impose any specific direction on the user, giving them the freedom to choose what samples to label next. However, we hypothesized that this freedom could be preserved while also assisting users by intermittently recommending a batch of examples for annotation. To test this hypothesis, we designed an experiment using a multi-armed bandit formulation, which recommended one of three actions to the user: the VA process, annotating random samples, or annotating zero-shot samples.

Each source is divided into batches of fixed size $d$. At each step $t$, an algorithm determines one source for annotation. We emulate the presentation of a batch of $d$ clips to an annotator and record the absolute gain in model quality compared to the previous step. At each time step $t$, the following occur:

\begin{enumerate}
    \item \textit{Selection}: The algorithm picks a source or "action", denoted as $i$. The choice of source depends on the bandit algorithm and the score associated with each source, which we denote as $s_i$.
    \item \textit{Action}: We add the next batch from the chosen source to the data from the previous step, which we denote as $D_{t-1}$. This results in a new dataset, $D_t$, on which we train a model (using the same setup as in the previous experiment). We then record the median test Average Precision (AP) as $v_t$.
    \item \textit{Update}: We update the per-source score for the chosen source, setting $s_i$ equal to the difference between the current $v_t$ and the most recent value $v_{t-1}$. We also ran experiments using simple and exponentially-weighted averages over the sequence of scores for each source, but found that they were inferior to using the most recent value.
\end{enumerate}

\begin{figure}
    \centering
    \includegraphics[width=0.9\linewidth]{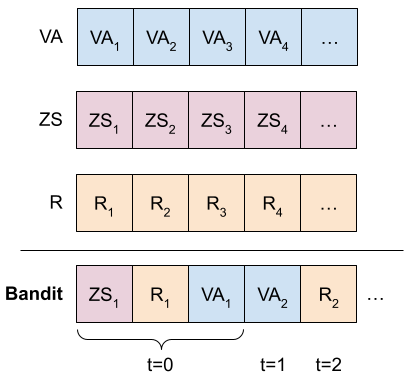}
    \caption[]{
    We first divide the data for each source into batches of size $d$, illustrated here as $\text{VA}_i$, $\text{ZS}_i$ (zero shot), and $R_i$ (random). At $t=0$, we independently evaluate the first batch for each source and compute $s_i$. At $t=1$, the algorithm chooses VA, prompting us to concatenate the first three batches with a new VA batch of size $d$ (i.e., $\text{VA}_2$). At $t=2$, the algorithm selects $R$, which leads to appending $R_2$.
    }
    \label{fig:bandit}
\end{figure}

We determine the initial score for each source, $s_i$, by evaluating the first batch for each source independently. For the first time step $t=1$, we set $D_0$ to the concatenation of the first batch from all three sources. An illustration of this process can be found in Fig \ref{fig:bandit}.

\subsubsection{Algorithms}
We implement the following algorithms:

\begin{enumerate}
    \item \textit{Round robin}: Serving as a baseline, this algorithm simply cycles through the three actions without referring to $s_i$.
    \item \textit{$\epsilon$-greedy}: With a probability of $\epsilon$, this algorithm randomly selects an action, and with a probability of $1-\epsilon$, it greedily selects the best action (i.e., highest $s_i$). We explored a few values of $\epsilon \in [0.1, 0.5]$ and found that $\epsilon=0.25$ works best.
    \item \textit{Upper Confidence Bound (UCB)} \cite{auer2002using}: This algorithm chooses the source with the highest value of $s_i + c \sqrt{\ln{N} / n_i}$, where $N$ is the total number of steps executed so far, and $n_i$ is the number of times arm $i$ has been selected before the current step. We tested various $c \in [10^{-3}, 1]$ and found $c=10^{-2}$ to perform the best.
    \item \textit{Greedy oracle}: Intended as a greedy upper bound, this algorithm evaluates all three actions at each step and selects the one yielding the highest $s_i$. Note that this is an oracle method and a priori knowledge of $s_i$ is not possible in practice.
\end{enumerate}

\subsubsection{Experimental setup}

The design of VA enables users to halt the labeling process once certain minimum conditions are satisfied (see Section \ref{methodology}). We used the final set of annotations and the corresponding model for each label (i.e. the largest value of $n$), which varies across labels.

We experimented with various values of batch size $d$ and set it to $d=25$ for this experiment, which resulted in a maximum of 40 steps and achieved the highest performance. Upon replicating these experiments with a larger batch size of $d=50$, we observed very similar outcomes. However, both larger ($d \ge 100$) and smaller ($d \le 10$) batch sizes led to a decline in performance.

\subsubsection{Results}

For each label, we calculate the $AP$ gain (similar to Section \ref{model_quality}) compared to VA without an algorithm for candidate source selection as the baseline. Table \ref{table:exp2} summarizes the distribution for each algorithm. We find that the median $AP$ gain is positive for all algorithms. UCB achieves the highest performance of 3.4 points (excluding the greedy oracle), resulting in a cumulative median improvement of 8.3 points over the most competitive baseline method CC (see Section \ref{exp1}). Interestingly, even a basic round robin approach can improve performance over the baseline.





\begin{table}[htbp]
\centering
\begin{tabular}{@{}llllll@{}}
\toprule
\multicolumn{1}{c}{\textbf{Method}} & \multicolumn{1}{c}{\textbf{$p_{10}$}} & \multicolumn{1}{c}{\textbf{$p_{25}$}} & \multicolumn{1}{c}{\textbf{$p_{50}$}} & \multicolumn{1}{c}{\textbf{$p_{75}$}} & \textbf{$p_{90}$} \\ \midrule
Round robin & -5.9 & -0.3 & 2.5 & 7.4 & 15.1 \\ \midrule
$\epsilon$-greedy & -6.5 & -0.6 & 2.9 & 7.6 & 16.7 \\ \midrule
UCB & -7.5 &  -0.2 & 3.4 & 9.2 & 15.9 \\ \bottomrule
Greedy oracle & -5.4 & 0.9 & 3.9 & 9.6 & 16.2          \\ \midrule
\end{tabular}
\caption{Distribution of model quality gain when using source selection relative to VA without an algorithm for candidate source selection. Column $p_n$ represents the $n$-th percentile, where $p_{50}$ is the median.}
\label{table:exp2}
\end{table}



\section{Limitations and future work}

We introduced Video Annotator (VA) as a framework for efficiently constructing binary video classifiers. There is a vast space of design decisions that can be explored in future work. 

At a high level, the framework's applicability extends beyond video to other media assets such as audio, images, text, and 3D objects. It can leverage advancements in joint embeddings across multiple modalities, such as ImageBind \cite{girdhar2023imagebind}. While we primarily explored context-free labels in this work, we could also extend this to long-form video by incorporating contextual encoders.

We can try to improve model performance in many ways, for example a superior video encoder could improve both sample efficiency and the final model's quality. Leveraging multimodal representations (e.g., audio and subtitles) might enhance performance for certain tasks, and advanced active learning strategies could further augment sample efficiency.


In terms of user experience, VA could be improved by refining metrics (model and data scores), proactively identifying potential issues (e.g. where the model's prediction and the users label do not agree), and suggesting additional search terms \cite{maniparambil2023enhancing}. More advanced searches, such as textual combined with metadata (e.g., genre, year), or video-to-video searches (i.e., identifying videos similar to a reference video) could also provide benefits.

\section{Conclusion}

We presented Video Annotator (VA), a novel, interactive framework that addresses many challenges associated with conventional techniques for training machine learning classifiers. VA leverages the zero-shot capabilities of large vision-language models and active learning techniques to enhance sample efficiency and reduce costs. It offers a unique approach to annotating, managing, and iterating on video classification datasets, emphasizing the direct involvement of domain experts in a human-in-the-loop system. By enabling these users to rapidly make informed decisions on hard samples during the annotation process, VA increases the system's overall efficiency. Moreover, it allows for a continuous annotation process, allowing users to swiftly deploy models, monitor their quality in production, and rapidly fix any edge cases. This self-service architecture empowers domain experts to make improvements without the active involvement of data scientists or third-party annotators, and fosters a sense of ownership, thereby building trust in the system.

We conducted experiments to study the performance of VA, and found that it yields a median 8.3 point improvement in Average Precision relative to the most competitive baseline across a wide-ranging assortment of video understanding tasks. We release a dataset with 153k labels across 56 video understanding tasks annotated by three professional video editors using VA, and also release code to replicate our experiments.

\section{Acknowledgements}
We'd like to thank Kelli Griggs, Eugene Lok, Yvonne Jukes, Anna Pulido, Alex Alonso, and Madeline Riley for valuable discussions and feedback.

\bibliographystyle{ACM-Reference-Format}
\bibliography{main}

\end{document}